\documentclass[conference]{IEEEtran}
\IEEEoverridecommandlockouts
\usepackage{cite}
\usepackage{amsmath,amssymb,amsfonts}
\usepackage{algorithmic}
\usepackage{graphicx}
\usepackage{textcomp}
\usepackage{xcolor}

\DeclareMathOperator*{\argmax}{argmax} 
\usepackage{makecell}
\usepackage{booktabs}
\usepackage{multirow}

\usepackage{footnotebackref}
\makeatletter
\LetLtxMacro{\BHFN@Old@footnotemark}{\@footnotemark}

\renewcommand*{\@footnotemark}{%
    \refstepcounter{BackrefHyperFootnoteCounter}%
    \xdef\BackrefFootnoteTag{bhfn:\theBackrefHyperFootnoteCounter}%
    \label{\BackrefFootnoteTag}%
    \BHFN@Old@footnotemark
}
\makeatother

\def\BibTeX{{\rm B\kern-.05em{\sc i\kern-.025em b}\kern-.08em
    T\kern-.1667em\lower.7ex\hbox{E}\kern-.125emX}}
\begin{document}

\title{Performance Improvement in Multi-class Classification via Automated Hierarchy Generation and Exploitation through Extended LCPN Schemes\\
}

\author{\IEEEauthorblockN{Celal Alagoz}
\IEEEauthorblockA{\textit{Electrical and Electronics Engineering} \\
\textit{Kırıkkale University}\\
Kırıkkale, Türkiye \\
0000-0001-9812-1473}}
\maketitle

\begin{abstract}
Hierarchical classification (HC) plays a pivotal role in multi-class classification tasks, where objects are organized into a hierarchical structure. This study explores the performance of HC through a comprehensive analysis that encompasses both hierarchy generation and hierarchy exploitation. This analysis is particularly relevant in scenarios where a predefined hierarchy structure is not readily accessible. Notably, two novel hierarchy exploitation schemes, LCPN+ and LCPN+F, which extend the capabilities of LCPN and combine the strengths of global and local classification, have been introduced and evaluated alongside existing methods. The findings reveal the consistent superiority of LCPN+F, which outperforms other schemes across various datasets and scenarios. Moreover, this research emphasizes not only effectiveness but also efficiency, as LCPN+ and LCPN+F maintain runtime performance comparable to Flat Classification (FC). Additionally, this study underscores the importance of selecting the right hierarchy exploitation scheme to maximize classification performance. This work extends our understanding of HC and establishes a benchmark for future research, fostering advancements in multi-class classification methodologies.
\end{abstract}

\begin{IEEEkeywords}
Hierarchical Classification, Multi-class Classification, Hierarchical Divisive Clustering, LCPN
\end{IEEEkeywords}

\section{Introduction}
In various domains such as text classification \cite{04klimt,05punera}, image understanding \cite{03shen}, annotation \cite{11dimitrovski,12dimitrovski}, and bioinformatics, particularly in the realm of protein function prediction \cite{03clare,06barut,09silla,09bianchi}, HC has emerged as a powerful approach. HC involves the organization of data or objects into a tree-like structure with nested categories or groups, establishing a hierarchy where each category is a subset of a larger one. This hierarchical structure offers advantages in handling complex datasets with structured labels.

Algorithms have been devised to harness this hierarchical structure for enhanced classification accuracy in cases where structured labels are readily available. However, in the majority of multi-class classification problems, structured labels are unavailable, and the potential performance benefits observed in hierarchical problems warrant further investigation. Only a limited number of studies have explored the advantages of inducing hierarchies from datasets typically associated with flat labels.

For instance, in a web content analysis study \cite{05punera}, an automatic taxonomy of documents was derived without predefined labels through the construction of a binary tree. The tree construction involved hierarchical clustering of classes using a top-down approach, employing Spherical K-means for cluster splitting. Subsequently, in a follow-up study \cite{06punera}, the superiority of n-ary trees over binary trees was reported in terms of classification performance.

In another investigation, researchers assumed the presence of a latent hierarchy in synthetic and various image data, resulting in a significant improvement in top-down classification tasks \cite{21helm}. They explored two distinct clustering methods for constructing hierarchies. The first method involved estimating the conditional means and clustering them using Gaussian Mixture Models (GMM), while the second method measured pairwise task similarity between conditional distributions and utilized a combination of spectral embedding and GMM for clustering. These innovative approaches shed light on the potential of hierarchical structures in classification scenarios where structured labels are not readily available, opening up new avenues for research in this domain.

The application of HC tasks varies based on several criteria \cite{11silla,01sun}. The first criterion concerns the type of hierarchical structure utilized, which is typically either a tree or a directed acyclic graph, allowing nodes to have multiple parent nodes.

The second criterion relates to the depth of classification in the hierarchy. This can involve either always performing classification down to the leaf nodes, known as mandatory leaf node prediction, or stopping at any node level, known as non-mandatory leaf node prediction.

The third criterion pertains to how the classifier is deployed within the hierarchical structure. Two approaches are commonly used: the global (or big-bang) classification, where a single classifier considers the entire class hierarchy, but lacks modularity, and the local classification, where local classifiers are placed at different regions of the hierarchical structure.

Local classifiers can be situated on each node (local classifier per node - LCN), at each level (local classifier per level - LCL), or on parent nodes (local classifier per parent node - LCPN). While different during training, these local classifiers share a similar top-down approach during the testing phase. In the top-down approach, more general and coarse labels are predicted at higher levels, while more specific and fine labels are predicted at lower levels as choices are narrowed down moving downward. Notably, the prediction of lower-level classes depends on the higher-level predictions, meaning misclassifications at higher levels propagate down the hierarchy.

In this study, the process of automatic hierarchy generation is executed by applying hierarchical clustering to class conditional means. In terms of modifying and extending the LCPN scheme, two strategies have been employed. Firstly, LCPN+ was developed to prevent predictions from occurring exclusively along a single path in the hierarchy. Additionally, LCPN+F was introduced to harmonize the LCPN and FC methods. This choice was made with the aim of striking a balance between the lack of modularity observed in global classification and the error propagation typically associated with local classification. The overarching objective of this approach is to optimize classification performance in multi-class datasets.

This study presents several noteworthy contributions to the field of hierarchical multi-class classification:

Novel Hierarchy Exploitation Schemes: This study introduces two innovative hierarchy exploitation schemes, LCPN+ and LCPN+F, which significantly contribute to the field of HC. These novel schemes offer a fresh perspective and advanced techniques for improving multi-class classification performance.

Consistent Superiority of LCPN+F: The research highlights the consistent superior performance of the LCPN+F scheme compared to other hierarchy exploitation schemes. This contribution provides a valuable solution for enhancing HC across various datasets and scenarios.

Balancing Modularity and Error Propagation: LCPN+F, in particular, strikes a balance between the modularity of global classification and the error propagation associated with local classification. This approach enhances classification performance in multi-class datasets, providing a valuable contribution to the field.

Efficiency in Classification: The study not only emphasizes the effectiveness but also the efficiency of the introduced schemes. LCPN+ and LCPN+F maintain runtime performance comparable to FC, making them practical choices for real-world applications.

Novel Hierarchy Generation Approach: In this study, an automatic hierarchy generation approach is introduced, which involves the utilization of hierarchical clustering applied to class conditional means. Furthermore, the study highlights the advantages of employing dimension reduction before obtaining class conditional means. This methodology offers a fresh perspective for creating hierarchies when predefined structures are unavailable.

Informed Decision-Making: The research underscores the importance of informed decision-making in selecting hierarchy generation and exploitation configurations. It provides insights into the relationship between dataset characteristics and suitable hierarchy generation techniques, aiding researchers and practitioners in making better choices for their specific tasks.

Extended Understanding of HC: By delving into the intricate interplay between hierarchy generation and exploitation, this work contributes to a deeper understanding of HC. It offers valuable insights that will guide future research in the domain, fostering advancements in multi-class classification methodologies.

Benchmark for Further Research: This study establishes a benchmark for evaluating and comparing hierarchy exploitation schemes, paving the way for further research and experimentation in HC. It sets a standard for assessing the performance of new techniques and approaches in this field.

In summary, this study's contributions offer insights into addressing the challenges of hierarchical multi-class classification, introducing a novel approach, and validating its efficacy through comprehensive experimentation. The findings presented here expand the understanding of HC methods and pave the way for further advancements in this evolving field. Open source code\footnotemark{} is available for the interested.

\footnotetext{\url{https://github.com/alagoz/hge_extended_lcpn}}

\section{Methods}
This study is structured into two distinct phases to enhance the performance of multi-class classification: hierarchy generation and hierarchy exploitation that leverages extended LCPN schemes. The subsequent sections provide in-depth discussions of these phases and detailed explanations of the classifier selection and evaluation methodologies.

\begin{figure*}[h]
\centerline{\includegraphics{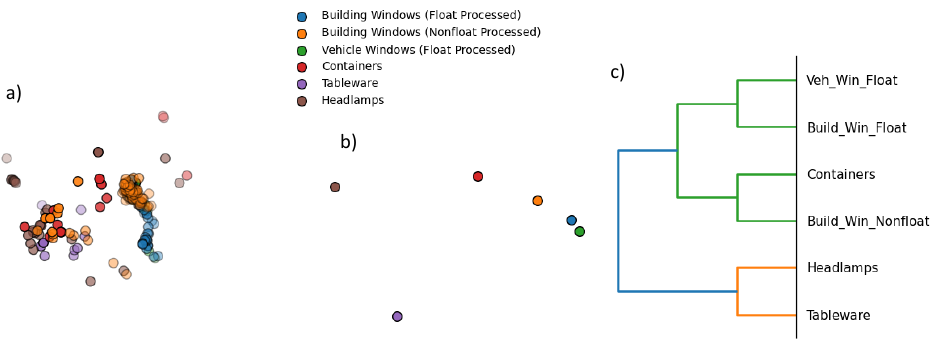}}
\caption{Demonstration of hierarchy generation via hierarchical divisive clustering using k-medoids as the clustering method on the Glass types dataset. (a) Data points are initially represented in a lower-dimensional 3D space and visualized as a point cloud. (b) Each class is subsequently represented as a point in the attribute space by computing class conditional means; in this case, there are six points corresponding to the six flat classes. (c) The final hierarchical structure is depicted as a binary tree after processing the class points using k-medoids.}
\label{fig:higen}
\end{figure*}
\subsection{Hierarchy Generation}
In HC, it is a common convention for the dataset to include a pre-defined hierarchy. Typically, this hierarchy is constructed based on intuitive, semantic, or ontological taxonomies, which provide a clear structure for organizing the classes. However, when a dataset only provides flat labels without any predefined hierarchy, the task of deriving a meaningful hierarchy can be quite challenging. Relying on intuition or semantics to construct a hierarchy may not necessarily lead to improvements in classification performance.

In such cases, a more pragmatic approach involves considering the attributes inherent to the flat class labels, as learned by the algorithms during training. Consequently, in this study, the challenge of hierarchy generation is addressed by treating each flat class label as an independent object and applying clustering techniques to organize them into a hierarchy. 

\paragraph{Class Conditional Means} Representing flat classes as objects and deriving their attributes can be achieved through the computation of class conditional means using the dataset. Class conditional means denote the average values of the features (attributes) within each class. Formally, given a dataset $D$ comprising $n$ instances and $m$ features, where each instance $i$ is represented as $x^{(i)} = (x^{(i1)},x^{(i2)},...,x^{(im)})$, and a set of $c$ flat classes $\{C_1,C_2,...,C_c\}$, the class conditional means for each class $C_j, \, (1 \leq j \leq c)$ can be calculated as follows:
\begin{equation}
\mu(C_j)=(\mu(C_j^1),\mu(C_j^2),...,\mu(C_j^m)) \label{eq1}
\end{equation}

Here, $\mu(C_j^i)$ represents the mean value of the j-th feature across all instances belonging to class $C_j$. Consequently, all data points associated with a particular class, represented as vectors in an m-dimensional space, are consolidated into a single point within that space. Within the context of hierarchical divisive clustering, these class conditional means serve as the input data for the clustering algorithm. 

\paragraph{Linear Discriminant Analysis} In this specific context, the application of dimension reduction before computing class conditional means serves as a valuable strategy to enhance the quality of hierarchical clustering. This approach focuses on extracting the most discriminative features while reducing the impact of noise in the data. To accomplish this, Linear Discriminant Analysis (LDA) is chosen as the dimension reduction technique.

LDA, as originally proposed by Fisher (1936) \cite{36fisher}, is a dimensionality reduction method with a primary objective of reducing the dimensionality of a dataset while simultaneously preserving the discriminative information between different classes. In essence, LDA seeks to identify a new set of dimensions, represented as linear combinations of the original features, in a manner that maximizes the variance between classes while minimizing the variance within each class. A key advantage of LDA is its supervised nature, which means it takes class labels into consideration during its dimension reduction process. By focusing on the class-related variance, LDA can reveal the most relevant features for distinguishing between classes, contributing to more effective clustering and classification results.

Dimension reduction, in this context, involves transforming data from the high-dimensional space $\mathbb{R}^{m}$ into a lower-dimensional space $\mathbb{R}^{k}$. The selection of the appropriate value for $k$ is implemented as an automated process. Prior to performing the actual dimension reduction, the dimension reduction algorithm used is trained using all $m$ components. This training process yields the percentage of variance explained by each of the components. Consequently, the sum of the explained variances across all components equals 1.0.

The selection of $k$ is determined by identifying the point at which the cumulative sum of variances surpasses or equals 0.95. In other words, $k$ is chosen such that the sum of the variances of the first $k$ components is at least 0.95. This method ensures that a significant portion of the variance in the data is retained in the lower-dimensional representation, helping to balance dimensionality reduction with the preservation of information.

\paragraph{Hierarchical Clustering} When generating a hierarchy through clustering, two distinct approaches can be applied: agglomerative and divisive clustering. Agglomerative clustering, often referred to as a bottom-up approach, initiates with individual objects considered as separate clusters, progressively merging them as the hierarchy is built from the bottom to the top. In contrast, divisive clustering begins with a single cluster encompassing all objects and divides it as the hierarchy develops from the top to the bottom. In both approaches, the clustering process relies on assessing the (dis)similarity between objects. To enable a comprehensive comparative analysis, both agglomerative and divisive clustering methods are implemented in this study.

In the case of hierarchical divisive clustering, the input to the clustering algorithm is provided by the class conditional means. For this purpose, the k-medoids algorithm \cite{05kaufman} is chosen as the clustering method. K-medoids, also known as partitioning around medoids, represents a variation of the k-means algorithm. In contrast to k-means, where cluster centers are computed as the averages of data points, k-medoids selects actual data points as cluster centers (referred to as medoids or exemplars). This characteristic enhances the interpretability of cluster centers, as they correspond to real data points within the dataset. Additionally, k-medoids demonstrates robustness against noise and outliers, making it a suitable choice for data with irregularities. While the k-medoids algorithm has the flexibility to work with a variety of dissimilarity measures, this study specifically opts for the Euclidean distance metric.

Conversely, in the hierarchical agglomerative clustering approach, the class conditional means also serve as input data. These means are utilized with the linkage function from the SciPy library, which is capable of handling either a distance matrix or observation vectors as inputs. In the present study, observation vectors are provided, and a pairwise distance matrix is constructed using the Euclidean distance metric. This distance matrix facilitates the hierarchical agglomerative clustering process, enabling the linkage function to create the hierarchical structure based on the provided dissimilarity information.

Figure \ref{fig:higen} illustrates the hierarchy generation process for the Glass dataset. In this case, dimension reduction is performed prior to computing class conditional means, followed by hierarchical divisive clustering. The resulting hierarchical structure takes the form of a binary tree. It's important to emphasize that this hierarchy structure is generated exclusively using the training set.
\begin{figure*}[h]
\centerline{\includegraphics{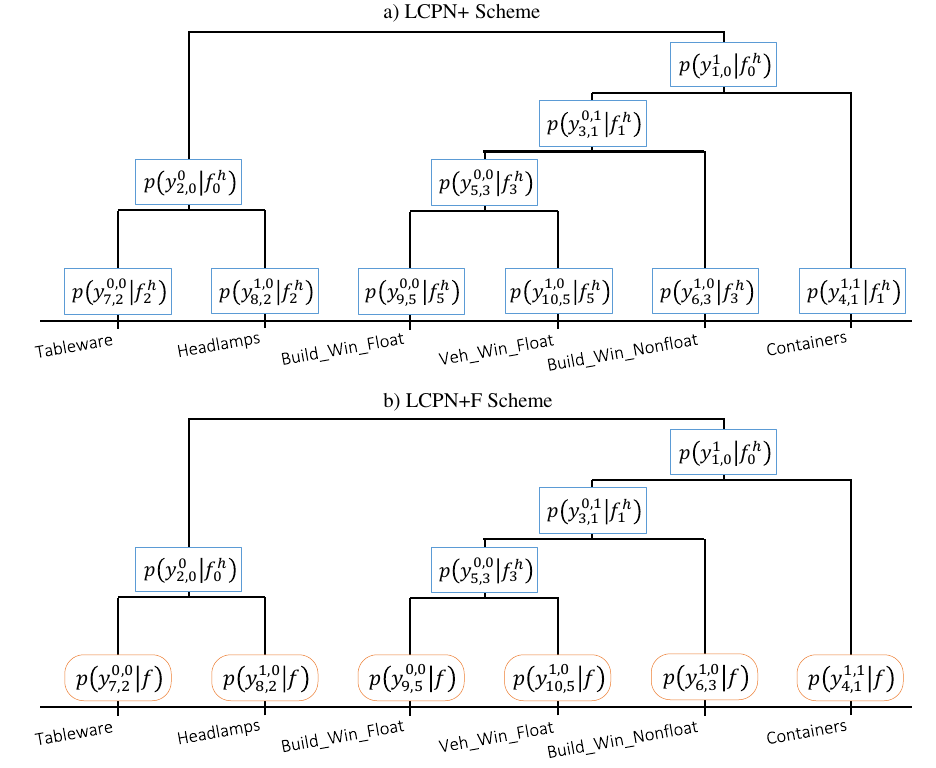}}
\caption{Hierarchy exploitation using extended LCPN schemes LCPN+ (a) and LCPN+F (b). Prediction in terms of conditional probabilities for each leaf (flat) and non-leaf node are displayed within individual boxes, including the node labels that are annotated based on the hierarchical structure and the classifier used for the prediction of the labels. Predictions for each label are made probabilistically. In LCPN, hierarchical classifiers predict all labels, whereas in LCPN+F, only leaf nodes are predicted by hierarchical classifiers, with the remaining labels predicted by the flat classifier. In the LCPN+F scheme in this specific hierarchy, hierarchical classifiers $f^h_2$ and $f^h_5$ are deactivated since they solely have leaf nodes as children.}
\label{fig:elcpn}
\end{figure*}

\subsection{Hierarchy Exploitation via Extended LCPN Schemes}
Enhancing the quality of the tree structure is a crucial step in addressing multi-class classification enhancement problems. Another critical phase is effectively utilizing this hierarchy, which involves deploying the classifier algorithm within the hierarchy structure. Depending on how the labels are considered by the classifier, two primary classification approaches emerge: global and local variants. In the global classification scheme, all labels are simultaneously considered by the classifier. This approach offers the advantage of internalizing attributes provided by the hierarchy in an encompassing way. However, it may exhibit weaknesses in handling intricate details and maintaining modularity. In contrast, local variants employ the classifier at each node, level, or parent node in the hierarchy. These schemes preserve modularity and internalize more details by leveraging local information. However, especially during the top-down prediction phase, misclassifications at higher levels can propagate errors affecting predictions at lower levels. This phenomenon is known as error propagation.

To address these issues and combine the strengths of both global and local classification approaches, this study introduces novel algorithms that extend the LCPN scheme. These algorithms are named LCPN+ and LCPN+F.

Before delving into the details of these algorithms, let us establish some notations. For a dataset with $c$ classes, there will be a total of $2c-1$ nodes, including leaf (singleton) and non-leaf (non-singleton) nodes within the hierarchy. Each node is assigned a label within the hierarchy, denoted as $y_{p,q}^{r,s}$, characterized by four subscripts – two subscripts and two superscripts. The first subscript $p, (0 \leq p \leq 2c-2)$ corresponds to the current node’s index, and the second subscript corresponds to the parent node’s index of the current node. The superscripts are employed to describe child indices – given the binary tree structure, child indices can be either 0 or 1, where the former represents the left child, and the latter represents the right child. The first superscript represents the child index of the current node, while the second superscript represents the child index of the parent node of the current node.

With these notations, $y_0$ denotes the root node, without superscripts since it has no child index, and without the second subscript since it has no parent node. Additionally, $f$ and $f^h_i$ represent the flat and hierarchical classifier functions, respectively. The subscript for the hierarchical classifier $f^h_i$ denotes the parent node index where the classifier is located. Lastly, $p(y|f)$ represents the probability of label $y$ predicted by the classifier function $f$. Figure \ref{fig:elcpn} illustrates a label setting in both the LCPN and LCPN+F hierarchical configurations.

\paragraph{LCPN+ Scheme} The LCPN+ scheme introduces a modification specifically for the prediction phase, while retaining the conventional LCPN approach during training. In the training phase, similar to conventional LCPN, only the classifiers at parent nodes are trained.

However, during the prediction phase, a departure from the conventional setting occurs. In the standard LCPN, an instance follows a single path of prediction. For example, if a certain instance is predicted at the root node for the right child label, it restricts the instance from being labeled with any labels falling under the left child label. This effectively narrows down the choices to a single flat label for the instance.

In LCPN+, labeling is carried out probabilistically, and all instances are predicted by all the classifiers. The prediction of a leaf node is then determined by considering the cascaded probabilities of that leaf node along with any connected non-leaf nodes. The classification of an instance is determined using the argmax operation, which selects the label with the highest predicted probability. This is formally defined as:

\begin{equation}
\argmax_{C_i \in C} \prod_{p,q,r,s \, \in \, \Omega_{C_i}} p(y_{p,q}^{r,s}|f^h_q) \label{eq_lcpn+}
\end{equation}

Here, $C_i$ represents a flat label within the hierarchy from a set of flat labels denoted as $C$, and $\Omega_{C_i}$ represents the path of labels for $C_i$. This path includes $C_i$ itself and its parent labels. $\Omega_{C_i}$ denotes the path of $C_i$  and is a composite set encompassing the $p$, $q$, $r$, and $s$ sets. For example, for the label $y_{6,3}^{1,0}$ in Figure \ref{fig:elcpn} (a), the individual paths are designated as $p=\{6,3,1\}$, $q=\{3,1,0\}$, $r=\{1,0,1\}$, and $s=\{0,1\}$. Consequently, the complete path is represented as $\Omega_{C_6}=\{\{6,3,1\},\{3,1,0\},\{1,0,1\},\{0,1\}\}$.

\paragraph{LCPN+F Scheme} In the LCPN+F scheme, LCPN+ is further enhanced by combining flat classifiers with hierarchical ones within the structure. This configuration introduces modifications during both the training and testing phases.

In the conventional LCPN approach, classifiers are exclusively employed at parent nodes, which may contain both non-leaf and leaf (flat) labels. In the LCPN+F scheme, classifiers at the parent nodes are dedicated solely to classifying non-leaf labels, leaving leaf labels to be classified by flat classifiers. When a parent node contains both leaf and non-leaf labels, the classifier at that node is trained conventionally but used only for predicting the non-leaf label during testing, bypassing the leaf label. However, if a parent node has both of its children  as leaf, it is completely deactivated, neither used for training nor prediction. Consequently, in the training phase, only the classifiers that remain active are trained.

The prediction phase in the LCPN+F scheme is similar to LCPN+ with the main difference being that leaf label predictions are made by a multi-class flat classifier. Similar to LCPN+, predictions are made probabilistically, where the prediction for each leaf node is computed through the chain rule of conditional probabilities down to that leaf. The leaf with the highest probability is then predicted. This is formally defined as:

\begin{equation}
\argmax_{C_i \in C} \prod_{p'\cup t ,q'\cup u, r'\cup v , s'\cup w \, \in \, \Omega_{C_i}} p(y_{p',q'}^{r',s'}|f^h_q)p(y_{t,u}^{v,w}|f) \label{eq_lcpn+}
\end{equation}

Here, $t$, $u$, $v$, and $w$ represent the last elements of individual paths for $\Omega_{C_i}$, while $p'$, $q'$, $r'$, and $s'$ are the remaining elements for the respective paths. The last term in the probability chain always represents the probabilistic output of a flat classifier.

Both in LCPN+ and LCPN+F, error propagation, typically encountered in LCPN, is circumvented by not constraining test instances to follow a single path. Furthermore, the incorporation of local information addresses a limitation of the global classification approach, thus having potential to improve the overall performance of the scheme. Additionally, for both schemes, the modifications enable the prediction phase to be implemented in parallel, potentially resulting in more efficient computation. Furthermore, the LCPN+F scheme offers enhanced efficiency as some parent nodes can be deactivated based on the hierarchy's structure.

\subsection{Classifiers and Evaluation}
For each dataset, various classifiers are tested, and the best-performing classifier is selected for use in the experiments. Time Series Forest (\texttt{tsf}) \cite{13deng} is exclusively chosen for time series datasets. The classifiers and their abbreviations are as follows:

\par \noindent \texttt{xgb} Extreme Gradient Boosting \cite{01friedman} is a high-speed, accurate machine learning algorithm that combines multiple decision trees for robust predictions.
\par \noindent \texttt{rf} Random Forest \cite{98ho} is an ensemble learning method that constructs multiple decision trees to enhance accuracy and reduce overfitting.
\par \noindent \texttt{lda} LDA \cite{36fisher} is a technique used in both classification and dimensionality reduction, aiming to find a lower-dimensional representation of data while maximizing class separation.
\par \noindent \texttt{nb} Naive Bayes is a probabilistic classification algorithm based on Bayes' theorem, assuming feature conditional independence.
\par \noindent \texttt{tsf} Time Series Forest is a machine learning algorithm specifically designed for time series classification tasks. It is an ensemble method that leverages randomization and decision trees to make accurate predictions on time series data.

Before comparing performances, some hyperparameters of the selected models are fine-tuned based on cross-validation results using the GridSearchCV function from the scikit-learn library. Table \ref{tab:dsets} presents the datasets and classifiers forming the baseline models for each dataset with their tuned parameters.

\begin{table}[t]
\caption{Datasets and Classifiers}
\begin{center}
\begin{tabular}{llllcl}
\toprule
Datasets & Archive & $c$ & Classifier & FC & Hyperparameters \\
\hline
Glass& UCI & 6 & \texttt{rf} & 75.72 & \makecell{max$\_$depth: 10 \\ n$\_$estimators: 300 \\ random$\_$state: 0} \\
\hline
PPTW$^{\mathrm{a}}$ & UCR & 6 & \texttt{tsf} & 54.61 & \makecell{n$\_$estimators: 50 \\ random$\_$state: 0} \\
\hline
Yeast& UCI &10&\texttt{xgb}&54.24&\makecell{learning$\_$rate: 0.25 \\ n$\_$estimators: 5 \\ random$\_$state: 0} \\
\hline
Faces&sklearn&40&\texttt{lda}&98.45&\makecell{solver: svd} \\
\hline
FiftyWords&UCR& 50 & \texttt{tsf}&61.23&\makecell{n$\_$estimators: 50 \\ random$\_$state: 0} \\
\bottomrule
\multicolumn{4}{l}{$^{\mathrm{a}}$ProximalPhalanxTW}
\end{tabular}
\label{tab:dsets}
\end{center}
\end{table}

Performance assessment relies on the F1-score metric, utilizing macro averaging for multi-class classification. The evaluation involves 5-fold cross-validation. Since hierarchical clustering is supervised, the training set examples are used for hierarchy tree construction, potentially generating up to 5 distinct trees.

Performance improvement in using HC over FC is quantified through a metric called Learning Efficiency (LE) \cite{20vogelstein}. LE is calculated as the ratio of F1-scores obtained from HC to FC. When LE is greater than 1 (LE $>$ 1), it indicates an enhancement in classification performance when employing HC.

\section{Experiments}
The experiments are conducted utilizing five real-world datasets sourced from various databases such as UCR \cite{ucr}, UCI \cite{uci}, and scikit-learn \cite{sklearn}. Neither of these datasets is furnished with pre-defined hierarchical labels, rendering them well-suited for the purposes of this study. This section offers a concise overview of the datasets used in the research, followed by the presentation of the experimental results.

\subsection{Datasets}
\textbf{Glass Identification dataset} \cite{glass} contains information about various chemical properties of glass, and the task is to classify different types of glass based on these attributes. The data consists of 214 instances, where each instance represents a different type of glass. There are ten attributes provided for each instance, such as the refractive index, sodium oxide content, magnesium oxide content, aluminum oxide content, silicon oxide content, potassium oxide content, calcium oxide content, barium oxide content, iron oxide content, and the type of glass. The target variable in the Glass dataset is the type of glass, which can belong to one of seven classes representing different glass types. The classification task aims to predict the type of glass based on the given chemical composition attributes. 

\textbf{ProximalPhalanxTW dataset} \cite{ptw} contains time series data, specifically related to hand movement. The dataset records measurements of the position of the proximal phalanx of a human hand over time. The dataset consists of several numerical attributes, where each attribute represents a different aspect of the hand movement. The specific number of attributes and instances may vary depending on the version of the dataset. Researchers often use the ProximalPhalanxTW dataset for tasks related to time series classification and pattern recognition. It can be employed to develop and evaluate algorithms for recognizing and categorizing hand movements based on the recorded data.

\textbf{Yeast dataset} \cite{yeast} represents a challenging problem of predicting the cellular localization sites of proteins in yeast cells. The dataset contains information about 1,484 proteins from Saccharomyces cerevisiae (commonly known as baker's yeast). Each protein is described by eight attributes, which are numerical and represent certain features extracted from the protein sequence. The target variable in the Yeast dataset is the cellular localization site, and it can belong to one of ten classes, representing different subcellular compartments where the proteins are located within the yeast cells. The primary task with the Yeast dataset is to classify proteins into their respective localization sites based on the provided attributes. Due to its complexity and multiclass nature, this dataset is widely used for evaluating and comparing the performance of various classification algorithms.

\textbf{Faces dataset} \cite{faces} The Olivetti Faces dataset, sourced from AT\&T Laboratories Cambridge, includes 400 grayscale images portraying 40 unique individuals. Each person is represented by 10 distinct images, showcasing variations in lighting, facial expressions, and poses. These images have a resolution of 64x64 pixels and are commonly employed by researchers in computer vision and machine learning for tasks such as face recognition and facial expression analysis.

\textbf{FiftyWords dataset} \cite{fiftywords} consists of a collection of handwritten texts. It contains samples of 50 different words, each handwritten by multiple individuals. Each sample is represented as a time series of pen-tip coordinates. The dataset consists of numerical attributes that represent the (x, y) coordinates of the pen-tip as it moves across a writing surface. Each time series captures the trajectory of the pen for a specific word. Researchers often use the FiftyWords dataset for tasks related to time series classification and pattern recognition. It's a valuable resource for developing and evaluating algorithms for handwritten word recognition. 

\subsection{Results}
All datasets undergo a one-time shuffling process before implementing cross-validation, with the random state set to 0 to ensure consistent data point distribution across multiple runs. In addition to LDA, various dimension reduction techniques, including PCA, Neighborhood Components Analysis \cite{nca}, Isomap \cite{isomap}, and Locally Linear Embedding \cite{lle} are employed. Among these, LDA consistently delivers the best results.

For hierarchical clustering, different distance metrics other than Euclidean are experimented with for both the k-medoids and linkage functions. Surprisingly, the Euclidean distance metric consistently yields higher-quality hierarchical trees, enhancing HC performance. Additionally, apart from k-medoids, k-means and Gaussian Mixture Model (GMM) clustering methods are also explored. However, k-medoids predominantly produces superior hierarchies, leading to improved HC performance.

After these initial configurations, results are obtained by generating hierarchies following dimensionality reduction applied to the datasets. Clustering is performed using both divisive and agglomerative approaches.

\begin{table}[t]
\caption{LE Results when Dimension Reduced with LDA}
\begin{center}
\begin{tabular}{llcccc}
\toprule
&Datasets & Global & LCPN & LCPN+ & LCPN+F \\
\midrule
\parbox[t]{1mm}{\multirow{5}{*}{\rotatebox[origin=c]{90}{Divisive}}} & Glass& 0.862 & 0.877 & 0.905 & \textbf{1.007} \\
&ProximalPhalanxTW &0.900 & \textbf{1.119} & \textbf{1.079} & \textbf{1.029} \\
&Yeast&0.467 & 0.935 & 0.905 & 0.897  \\
&Faces&0.945 & 0.845 & 0.848 & \textbf{1.002}\\
&FiftyWords&0.660 & 0.842 & 0.927 & \textbf{1.022}\\
\midrule
\parbox[t]{1mm}{\multirow{5}{*}{\rotatebox[origin=c]{90}{Agglomerative}}} & Glass& 0.512 & 0.936 & 0.954 & 0.972\\
&ProximalPhalanxTW &0.918 & \textbf{1.064} & \textbf{1.063} & \textbf{1.091} \\
&Yeast&0.090 & \textbf{1.002} & 0.984 & 0.942  \\
&Faces&0.263 & 0.929 & 0.931 & \textbf{1.005}  \\
&FiftyWords&0.054 & 0.784 & 0.993 & 0.946  \\
\bottomrule
\end{tabular}
\label{tab:lda}
\end{center}
\end{table}

Hierarchy exploitation involves the utilization of four distinct schemes: global, LCPN, LCPN+, and LCPN+F, with each scheme being systematically evaluated. The LE results for each scheme are presented in Table \ref{tab:lda}, with cases of performance improvement over the baseline being highlighted in bold font. It is evident that the choice of hierarchy exploitation scheme has a discernible impact on the performance of HC.

Notably, the standout finding is the consistent superiority of the LCPN+F scheme in comparison to the other schemes. This trend is particularly pronounced when employing the LCPN+F scheme in scenarios involving hierarchical agglomerative clustering, where performance improvements over FC are observed across all datasets, except for the Yeast dataset. However, when hierarchical agglomerative clustering is used, suboptimal results are noted, especially in the cases of Yeast, Glass, and FiftyWords datasets. This underscores the suitability of the LCPN+F scheme for hierarchical agglomerative clustering during the hierarchy generation process.

\begin{table}[t]
\caption{LE Results when no Dimension Reduction Performed}
\begin{center}
\begin{tabular}{llcccc}
\toprule
&Datasets & Global & LCPN & LCPN+ & LCPN+F \\
\midrule
\parbox[t]{1mm}{\multirow{5}{*}{\rotatebox[origin=c]{90}{Divisive}}} & Glass& 0.910 & 0.837 & 0.887 & 0.984 \\
&ProximalPhalanxTW &\textbf{1.052} & \textbf{1.048} & \textbf{1.112} & \textbf{1.119} \\
&Yeast&0.528 & 0.804 & 0.810 & 0.890 \\
&Faces&0.946 & 0.747 & 0.757 & 0.995 \\
&FiftyWords&0.409 & 0.925 & 0.993 & \textbf{1.012}\\
\midrule
\parbox[t]{1mm}{\multirow{5}{*}{\rotatebox[origin=c]{90}{Agglomerative}}} & Glass& 0.672 & 0.922 & 0.910 & 0.984\\
&ProximalPhalanxTW &0.834 & \textbf{1.108} & \textbf{1.102} & \textbf{1.02} \\
&Yeast&0.124 & \textbf{1.041} & 0.996 & 0.938 \\
&Faces&0.278 & 0.907 & 0.910 & 0.997 \\
&FiftyWords&0.105 & 0.842 & \textbf{1.004} & 0.936  \\
\bottomrule
\end{tabular}
\label{tab:no_redu}
\end{center}
\end{table}

The second-best performing scheme is LCPN+, which consistently delivers modest enhancements over the LCPN scheme in most instances. Notably, exceptions to this pattern are observed in the ProximalPhalanxTW and Yeast datasets. Surprisingly, for both divisive and hierarchical agglomerative clustering, these two datasets exhibit a performance trend contrary to that observed in the other datasets when utilizing the LCPN, LCPN+, and LCPN+F schemes. In most other datasets, there is a progressive increase in performance as one moves from LCPN to LCPN+ and then to LCPN+F schemes, while LCPN achieving the best performance in the ProximalPhalanxTW and Yeast datasets.

Regarding global classification, erratic fluctuations in performance are observed, with notable declines, particularly in the case of hierarchical agglomerative clustering applied to datasets like Yeast, Faces and FiftyWords. Overall, global classification exhibits the poorest performance among the tested schemes. In a general sense, hierarchies generated through the hierarchical agglomerative clustering approach consistently outperform those generated using hierarchical agglomerative clustering. An exception to this trend is observed only in the case of LCPN, which achieves better results when employing hierarchies generated through hierarchical agglomerative clustering.

\begin{figure*}[t]
\centerline{\includegraphics{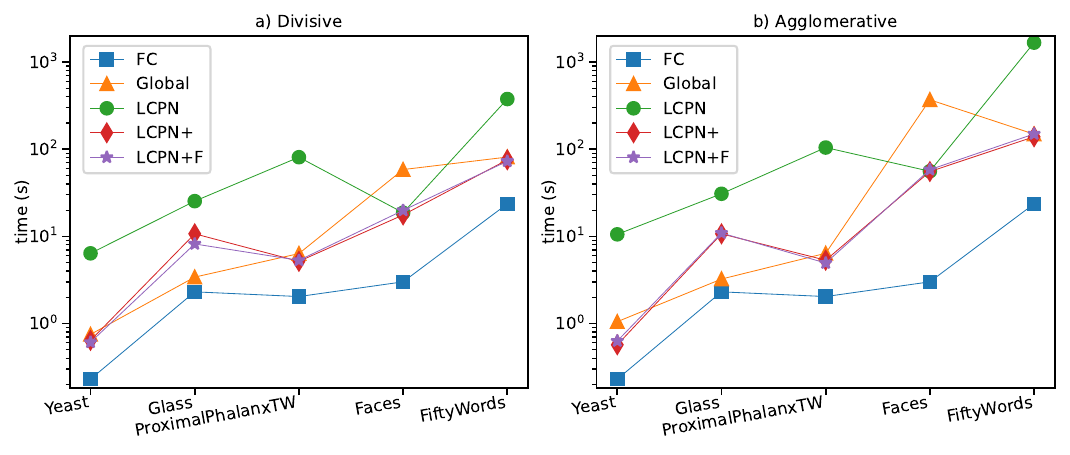}}
\caption{Empirical runtime of hierarchy exploitation schemes when hierarchy was generated using divisive (a) and agglomerative (b) hierarchical clustering approach. Speed of Global, LCPN+ and LCPN+F scehmes are comparable to FC whlie LCPN is the slowest among them.}
\label{fig:runtime}
\end{figure*}

Hierarchy exploitation schemes are further examined while generating hierarchies without applying dimension reduction to the data points during the hierarchy generation phase. The results of the LE technique are displayed in Table \ref{tab:no_redu}, with performance improvements over the baseline (FC) once again highlighted in bold font. In general, a decline in performance when comparing scenarios without dimension reduction is observed.

Remarkably, global classification surprisingly demonstrates significantly better performance in the ProximalPhalanxTW dataset than it typically exhibits in other cases. However, it still yields notably low results for the Yeast, Faces, and FiftyWords datasets when hierarchies are generated using hierarchical agglomerative clustering. Additionally, in the case of hierarchical divisive clustering, the behavior of the ProximalPhalanxTW and Yeast datasets differs from their performance when dimension reduction is applied. They exhibit the best performance with the LCPN+F scheme, with a progressive decline as moving to LCPN+ and then LCPN.

It is essential to highlight that the ProximalPhalanxTW dataset consistently shows performance improvement across all LCPN and its extended variants —LCPN+ and LCPN+F— across all cases, suggesting that certain datasets are particularly well-suited for HC. As an exception, an improvement is noted in the Yeast dataset when using LCPN for hierarchy exploitation and hierarchical agglomerative clustering for hierarchy generation. In contrast, LCPN+F exhibits more modest performance, achieving performance improvements over FC only in the ProximalPhalanxTW and FiftyWords datasets, both of which are time series datasets.

Meanwhile, when employing hierarchies generated through hierarchical agglomerative clustering, both time series datasets—ProximalPhalanxTW and FiftyWords—consistently demonstrate high performance, regardless of whether dimension reduction is applied. This observation provides valuable insights into the characteristics of time series datasets whether dimension reduction is employed or not before the hierarchy generation phase. However, further analysis of additional examples is required to elucidate these properties fully and arrive at more conclusive findings.

Lastly, empirical runtime analysis was conducted to assess the speed and efficiency of hierarchy generation and exploitation schemes. For the empirical computational cost analysis, the total running time of the 5-fold cross-validation procedure was considered. It is important to emphasize that the HC schemes encompassed an additional hierarchy generation step during the training phase, which inherently included dimension reduction. This hierarchy generation step was factored into the total running time calculations.

The empirical runtime analysis, as shown in Figure \ref{fig:runtime}, focuses on scenarios involving hierarchy trees generated through both hierarchical divisive clustering and hierarchical agglomerative clustering, in conjunction with dimension reduction. As anticipated, FC consistently emerged as the fastest-running algorithm in the experiments.

The LCPN+, LCPN+F, and global classification schemes exhibited runtime performance comparable to FC in both hierarchical divisive clustering and hierarchical agglomerative clustering scenarios. They operated at a pace approximately three times slower than FC. However, LCPN stands out as the slowest algorithm in this context, with runtime speeds approximately 100 times slower than FC.

Furthermore, it is worth noting that LCPN+ and LCPN+F consistently displayed very similar runtime performance, while global classification, in general, was observed to be slower. An exception to this trend was identified in the Glass dataset within the hierarchical agglomerative clustering context, where global classification exhibited a faster runtime.

\section{Discussions}
In this study, a comprehensive empirical analysis was conducted to examine various aspects of HC performance. Multiple experiments and considerations were performed to understand the factors influencing HC outcomes. In this discussion, the implications of the findings are discussed, considering aspects such as data preprocessing, dimensionality reduction, distance metrics, clustering methods, and hierarchy exploitation schemes.

\paragraph{Data Preprocessing and Dimensionality Reduction} The experiments began with a data preprocessing step involving a one-time shuffling process. A consistent data point distribution was maintained across multiple runs by setting the random state to 0. Additionally, different dimension reduction techniques, including PCA, NCA, Isomap, and LLE, were employed. Remarkably, LDA consistently outperformed other techniques, highlighting its effectiveness in feature extraction for HC.

\paragraph{Choice of Distance Metric and Clustering Method} For hierarchical clustering, various distance metrics beyond the Euclidean distance were explored, testing their suitability for both k-medoids and linkage functions. Surprisingly, the Euclidean distance metric consistently produced higher-quality hierarchical trees, resulting in improved HC performance. Additionally, different clustering methods, including k-means and GMM, were experimented with, but it was found that k-medoids predominantly generated superior hierarchies, leading to improved HC performance.

\paragraph{Hierarchy Exploitation Schemes} The analysis extended to hierarchy exploitation schemes, where four distinct schemes—global, LCPN, LCPN+, and LCPN+F—were systematically evaluated. The results indicated a discernible impact on HC performance based on the choice of hierarchy exploitation scheme. Notably, the consistent superiority of the LCPN+F scheme compared to the other schemes was observed. This trend was prevalent across most datasets, except for the Yeast dataset. Conversely, suboptimal results in certain cases, including Yeast, Glass, and FiftyWords datasets, were observed when hierarchical agglomerative clustering was used. These findings underscored the suitability of the LCPN+F scheme for hierarchical divisive clustering during hierarchy generation. The second-best performing scheme was LCPN+, consistently delivering modest enhancements over the LCPN scheme in most cases. However, exceptions were observed, notably in the ProximalPhalanxTW and Yeast datasets. Surprisingly, these two datasets exhibited performance trends contrary to those observed in other datasets when utilizing the LCPN, LCPN+, and LCPN+F schemes. In most other datasets, there was a progressive increase in performance as the transition was made from LCPN to LCPN+ and then to LCPN+F schemes, while LCPN achieved the best performance in the ProximalPhalanxTW and Yeast datasets. The analysis also touched upon global classification, which exhibited erratic fluctuations in performance, with notable declines, particularly in the case of hierarchical agglomerative clustering applied to datasets like Yeast, Faces, and FiftyWords. Overall, global classification demonstrated the poorest performance among the tested schemes.

\paragraph{Impact of Dimension Reduction} An analysis was conducted that excluded dimension reduction in the hierarchy generation phase. The results indicated a decline in performance when comparing scenarios without dimension reduction. Notably, global classification surprisingly demonstrated significantly better performance in the ProximalPhalanxTW dataset than in other cases, although it still yielded notably low results for the Yeast, Faces, and FiftyWords datasets when hierarchies were generated using hierarchical agglomerative clustering. The behavior of the ProximalPhalanxTW and Yeast datasets differed in the case of hierarchical divisive clustering without dimension reduction, with a preference for the LCPN+F scheme and a progressive decline as LCPN+ and LCPN were employed.

\paragraph{Time Series Datasets and Dimension Reduction} In the context of time series datasets, specifically the ProximalPhalanxTW and FiftyWords datasets, a notable observation emerged. It was consistently observed that the ProximalPhalanxTW dataset exhibited performance improvements across all LCPN variants—LCPN, LCPN+, and LCPN+F—across all cases. This suggests that certain datasets are inherently well-suited for HC, irrespective of the hierarchy exploitation scheme employed. Furthermore, in both scenarios, whether dimension reduction was applied or not, LCPN+F consistently demonstrated improvements over FC in the context of time series datasets. This observation hints at the potential versatility and effectiveness of the LCPN+F scheme in managing time series data, regardless of whether dimension reduction is integrated into the process. However, to validate and explore this finding further, additional time series datasets should be investigated.

\paragraph{Runtime Analysis} An empirical runtime analysis provided insights into the efficiency of hierarchy generation and exploitation schemes. Despite an additional hierarchy generation step during the training phase, the LCPN+, LCPN+F, and global classification schemes exhibited runtime performance comparable to FC in both hierarchical divisive clustering and hierarchical agglomerative clustering scenarios. They operated at approximately three times the runtime of FC.

In conclusion, a comprehensive analysis was conducted to understand the factors influencing HC performance. The findings contribute to informed decisions in designing HC systems tailored to specific datasets and applications.

\subsection{Future Directions}
The analyses conducted in this study have provided valuable insights into the impact of both hierarchy generation and hierarchy exploitation on HC performance. Consequently, the task of selecting the most appropriate configuration for both hierarchy generation and hierarchy exploitation for a given dataset emerges as a critical endeavor in achieving desired enhancements in multi-class classification. Specifically, establishing a deeper understanding of the relationship between dataset characteristics and suitable hierarchy generation techniques can facilitate more informed decision-making when choosing an appropriate technique. Furthermore, once a hierarchy is generated, it may possess unique properties that can inform the selection of the most effective exploitation scheme.

Another promising avenue for future research involves the exploration of additional exploitation schemes. Identifying the optimal configuration for a specific dataset necessitates a comprehensive analysis across multiple datasets. As a result, future work will encompass an expansion of this analysis to include a broader range of datasets, further enhancing the understanding of the intricate interplay between data characteristics, hierarchy generation techniques, and exploitation schemes in the realm of HC.

\section{Conclusions}
This study has delved deeply into the realm of HC, providing a comprehensive analysis of its performance. Notably, novel hierarchy exploitation schemes have been developed, specifically, LCPN+ and LCPN+F, marking significant advancements in this field.

Throughout the analyses conducted, LCPN+F emerged as the standout performer, consistently delivering exceptional results in terms of both classification efficiency and effectiveness.

The findings underscore the pivotal role of selecting the appropriate configuration for both hierarchy generation and exploitation in HC. These newly introduced schemes, LCPN+ and LCPN+F, have demonstrated their prowess in this context.

Additionally, the empirical runtime analysis has highlighted the efficiency of these schemes, with LCPN+ and LCPN+F performing commendably while maintaining runtime performance comparable to FC.

In summary, this study represents a substantial stride forward in the domain of HC, introducing novel schemes in the form of LCPN+ and LCPN+F, which exhibit superior performance and efficiency. These schemes hold immense promise for enhancing multi-class classification tasks, and the findings presented here provide valuable insights that will guide future research endeavors in this field.

\end{document}